\pdfoutput=1
\documentclass[runningheads]{llncs}
\let\paragraph\oldparagraph
\let\subparagraph\oldsubparagraph
\usepackage{graphicx}
\usepackage{caption} 
\usepackage{titlesec}

\begin{document}
\title{From Pivots to Graphs: Augmented Cycle Density as a Generalization to One Time Inverse Consultation \thanks{Extension to work done under the guidance of Dr. Jorge Gracia and Dr. Mikel L. Forcada. Supported by Apertium through the Google Summer of Code Program}}
%
%
\author{Shashwat Goel\inst{1} \and
Kunwar Shaanjeet Singh Grover\inst{1}} 
%
\authorrunning{S. Goel and K.S.S. Grover}
%
\institute{International Institute of Information Technology, Hyderabad, Telangana, India}
\maketitle              
\begin{abstract}
This paper describes an approach used to generate new translations using raw bilingual dictionaries as part of the $4^{th}$ Task Inference Across Dictionaries (TIAD 2021) shared task. We propose Augmented Cycle Density (ACD) as a framework that combines insights from two state of the art methods that require no sense information and parallel corpora: Cycle Density (CD) and One Time Inverse Consultation (OTIC). The task results show that across 3 unseen language pairs, ACD's predictions, has more than double (74\%) the coverage of OTIC at almost the same precision (76\%). ACD combines CD's scalability - leveraging rich multilingual graphs for better predictions, and OTIC's data efficiency - producing good results with the minimum possible resource of one pivot language.

\end{abstract}
\section{Introduction}
\subsection{Motivation}
Traditional \emph{bilingual} dictionaries provide a mapping between only a pair of languages. Their creation requires fluent speakers in both languages, which can be a difficult when both languages are uncommon. To incorporate translations between a wide variety of languages, the focus is gradually shifting to a graph-based representations of translations, such as the Linguistically Linked Open Data \cite{Cimiano2020LLD} project. Motivated by this graph structure, we analyze the 2 most popular automatic dictionary induction heuristics: Cycle Density (CD) \cite{Villegas2016} and One Time Inverse Consultation (OTIC) \cite{Tanaka1998}. We then propose a combined framework called \emph{Augmented Cycle Density} that can be viewed as a generalized version of OTIC beyond just one pivot. Both techniques leverage raw word (phrase) level translation data to either enrich existing language pairs or generate new language pairs altogether, which aligns perfectly with the objective of the Translation Inference Across Dictionaries (TIAD 2021) shared task. 

\subsection{Why Graph Heuristics?}
The Graph-based heuristics we study require only minimal existing language resources: dictionaries in their rawest form as pairs of words that form valid translations, without any sense information and parallel corpora. This makes them particularly suited to under-resourced languages. Infact, the ACD framework we propose can even produce good results with a single pivot language, reducing to OTIC in this setting. We emphasize some advantages of the graph-heuristic methods we study (CD, OTIC, ACD) over the class of methods based on distributional semantics:
\begin{itemize}
\item The graph heuristic methods do not need large corpora, which can be hard to obtain for under-resourced languages. Further, they do not require complex augmentation procedures.
\item Since antonyms often occur in similar contexts in sentences, they are incorrectly assigned high similarity by distributional semantics methods.
\item Distributional semantics based methods ignore polysemy. They are known \cite{Liu2018Homographs} to perform suboptimally when finding synonyms for polysemous words because they aggregate statistics across the different semantic senses \cite{Arora2018Linear}.
On the other hand, the graph heuristics we study are explicitly polysemy-aware.
\end{itemize}
Besides these advantages in prediction quality, graph based techniques are typically much more computationally efficient. This allows much faster tuning of hyperparameters and input data to produce the best results. Currently, they are also much more interpretable as the exact selection criteria is deterministic.

\subsection{Translation Graphs}
\emph{Translations} can be represented by an undirected graph $G(V, E)$, which we call a \emph{Translation Graph} $G_T$. Each vertex $ v \in V$ represents a lexical entry, such as the tuple: $\langle \mathrm{rep}, \mathrm{lang}, \mathrm{pos} \rangle$ where $\mathrm{rep}$ is the written representation of its canonical form, $\mathrm{lang}$ is the language, and $\mathrm{pos}$ is the part of speech (POS). Additional information can be stored as well. An edge $ e(u, v) \in E$ indicates that the lexical entries $u$ and $v$ are connected through a translation relation, therefore sharing at least one lexical sense. This definition \cite{Goel2021} is closely related to TransGraphs \cite{Mausam2009} in existing NLP literature and Semantic Maps in Cognitive Science.

One method to populate such a graph is by leveraging multiple bilingual dictionaries many of which are freely available online through projects like Apertium RDF \cite{Gracia2018RDF}. We adopt the same to stay consistent with the TIAD'21 test set distribution.

\begin{figure}
\centering
\includegraphics[width=0.4\textwidth]{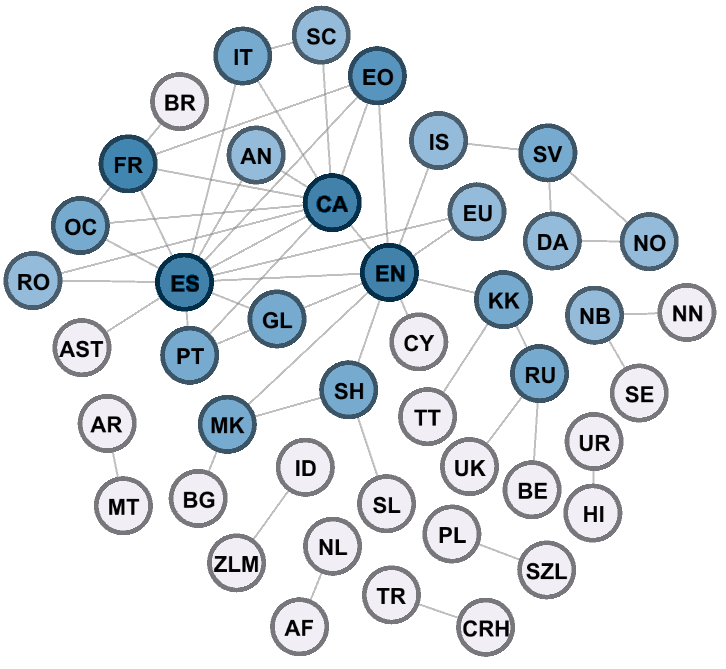}
\caption{Apertium RDF graph (figure taken from~\cite{Gracia2020leveraging}), which covers 44 languages and 53 language pairs. The nodes represent monolingual lexicons and edges the translation sets among them. Darker nodes correspond to more interconnected languages.}
\label{fig:ApertiumRDFv2.0}
\end{figure}

\subsection{System Summary}
Our system utilizes the improved open-source implementation\footnote{\url{https://github.com/shash42/ApertiumBidixGen/tree/master/Tool}} of the cycle density algorithm \cite{Villegas2016} released by \cite{Goel2021} as the ApertiumBidixGen tool. We retain the performance, tuning and evaluation recommendations of \cite{Goel2021}. We augment the cycle density algorithm with insights from OTIC, which leads to a system that can reduce to both OTIC and CD for specific hyperparameter choices. The augmentation has minimal impact on the efficiency of the system, producing entire new language pairs within 1 minute just like \cite{Goel2021}. The OTIC algorithm has consistently performed the best in all past editions of TIAD, but our system manages to improve results considerably.

\section{Method}
\subsection{One Time Inverse Consultation}
\label{sec:OTIC}
Let $(X, Y)$ represent a bilingual dictionary between two languages $X, Y$.
The OTIC algorithm, uses a pivot language $C$ to construct a dictionary $(A, B)$ given that the dictionaries $(A, C)$ and $(C, B)$ are already available.\\

The algorithm produces all translation pairs $a \in A$ and $b \in B$ which satisfy one of the following 2 types:

\begin{enumerate}
  \item \textbf{Type A}: The set of
  translations obtained by consulting $(A, C)$ for the word $a$ and the set of translations obtained by consulting $(C, B)$ for the word $b$ have at least two common translations.
  \item \textbf{Type B}: On consulting $a$ in
  dictionary $(A, C)$ and then consulting the obtained translations in
  dictionary $(C, B)$, we get only one final translation $b$. The rationale is that this probably means the mapping did not involve any polysemous words since it leads to only one possibility in the target language.
\end{enumerate}

\subsection{Cycle Density Algorithm}
The CD algorithm, instead of using a single pivot language, uses the entire translation graph to obtain possible translations between the languages $A$ and $B$. The algorithm is as follows:

\begin{enumerate}
    \item Given the translation graph $G_T$, for each word $a \in A$ compute all cycles containing $a$ that satisfy some constraints for performance and accuracy optimization. Currently studied constraints include conditions on minimum and maximum length, upper bounds on shortest distance of each word from $a$ in the cycle etc. Some non-polysemous POS like propernouns and numerals are translated transitively \cite{Goel2021}.
    \item The confidence score of a translation from $a$ to $b$ not directly
    connected in $G_T$ is the density of the densest cycle containing both $a$ and $b$.
    The density of a cycle is defined as the density of the subgraph induced by the cycle. The density of a subgraph $G^\prime(V^\prime, E^\prime)$ is defined as the ratio of actual number of edges to the number of edges that would be found in a fully connected graph (\emph{clique}): $\frac{2|E^\prime|}{|V^\prime|(|V^\prime| - 1)}$. \item All translations with confidence score above a chosen threshold are selected in the final prediction set.
\end{enumerate}

\subsection{Augmentation to the original CD algorithm based on OTIC}
Our modification to the CD algorithm involves augmenting it with Type B translations as defined in section~\ref{sec:OTIC}. That is, in addition to the CD predictions, we take all
translations $(a \in A, b \in B)$ such that the word $b$ is the only transitive mapping in language $B$ for the word $a$ of language $A$.

For this augmentation we only take a single pivot language. Taking a graph with 
multiple languages acting as intermediates results in much fewer words in source language $A$ having only one transitive correspondence in $B$.\\

With this augmentation, our final algorithm: \emph{Augmented Cycle Density} is as follows:
\begin{enumerate}
    \item \textbf{Type-A}: Run the CD algorithm and generate a list of possible translations and confidence scores in the range $[0, 1]$ using the entire graph.
    \item \textbf{Type-B}: Run the OTIC Type-B based augmentation separately which selects the word $b \in B$ as a translation for $a \in A$ if it is the only possible correspondence found transitively.  We empirically find such translations to have high precision and thus add them to the list of possible translations with confidence 1. 
    \item Choose the final prediction set from the list of possibilities by setting an appropriate confidence threshold.
\end{enumerate}

\subsection{ACD as a generalization to OTIC}
Notice that the Type A translations in OTIC are essentially all pairs of words which have atleast one cycle going through them. The CD algorithm is a generalization beyond one pivot, with the density threshold becoming important as with more pivots and longer intermediate paths, the chances of \emph{spurious cycles} which contain \emph{correlated polysemy} increases. Each additional edge increases the density and the chance that the cycle's induced subgraph should be a clique, and thus the words a valid translation pair.
 
The Augmented Cycle Density algorithm retains precisely the predictions of the OTIC algorithm if we choose only one pivot for the CD based generation and set the confidence threshold to \emph{non-zero}. For every source word, the target either shares atleast one cycle (OTIC type A) or is the only correspondence through the pivot (OTIC type B), giving the exact same prediction set.

In general, any improvements made to the cycle density algorithm can be leveraged in the Augmented Cycle Density framework for better Type A translations. This helps scale the algorithm with the availability of more bilingual dictionaries, something which OTIC cannot benefit from as it uses only one pivot. On the other hand, OTIC produces good results with the minimum resource of one pivot and is extremely data efficient \cite{Goel2021}, unlike CD which depends on richer graphs to find sufficient cycles. This makes OTIC especially useful for language pairs which aren't well connected (as long as there is a common pivot). By reducing to them under specific hyperparameter choices and generally combining their predictions, the ACD framework provides the best of both worlds: the data efficiency of OTIC and scalability of CD.

\subsection{Choices Specific to TIAD 2021 Requirements}
TIAD 2021 required pairwise generation of translations between English (en), French (fr) and Portuguese (pt). These language pairs are not available currently in Apertium RDF and resources pertaining to these pairs were not allowed in the task. 

For the CD step, we used as input a subgraph of 27 language pairs across 13 languages forming the largest biconnected component in the Apertium RDF graph, the same as the \emph{large set} in \cite{Goel2021}. We used the default hyperparameter settings recommended in \cite{Goel2021}: Transitivity at context depth 4 for the "numeral" and "proper noun" POS, and Context depth 3, minimum cycle length 4, maximum cycle length 6, confidence threshold 0.6 for all other POS. 

\begin{figure}
\centering
\includegraphics[width=0.48\textwidth]{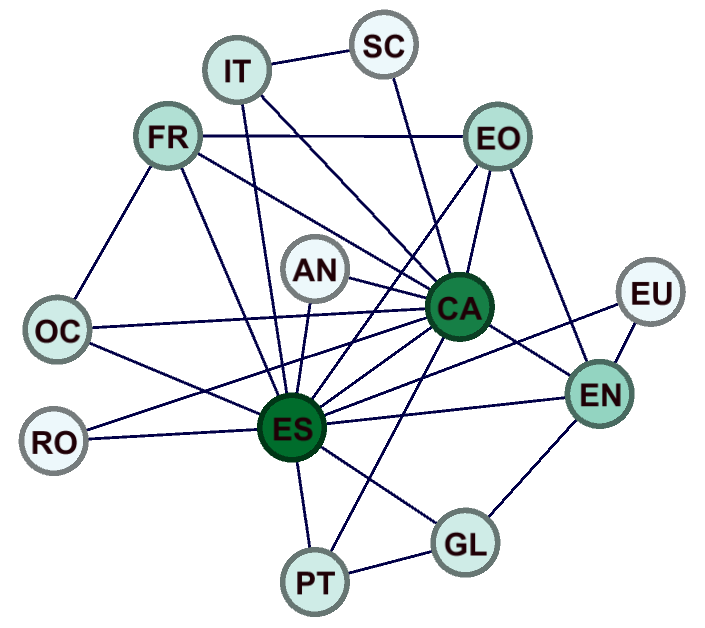}
\caption{The subset of the Apertium RDF graph used in the CD step of our translation generation procedure for TIAD} \label{fig:CDStepGraph}
\end{figure}

In the augmentation step we had multiple possible pivot choices for each pair. We choose Catalan as it is a common pivot across the 3 language pairs and has large bilingual dictionaries. While validating our choice, we found the coverage and precision to vary significantly based on the pivot chosen as some Apertium bilingual dictionaries are far richer than others.
\begin{table}
\centering
\begin{tabular}{|l|l|}
\hline
Pair &  Pivot Candidates\\
\hline
En-Fr & Ca, Eo, Es\\
\hline
En-Pt & Ca, Es, Gl\\
\hline
Fr-Pt & Ca, Es\\
\hline
\end{tabular}
\caption{Possible pivots for each test pair}\label{tab1}
\end{table}

To validate our procedure and observations before the TIAD submission, we benchmark on the task of generating an English-Catalan dictionary from scratch. This is because this pair's Apertium dictionary is rich and can be used as a reference. We focus on the Both Word Precision (BWP) and Both Word Recall (BWR) metrics as defined in \cite{Goel2021} to compare OTIC, CD and ACD, using Spanish as pivot throughout.

\section{Results}

In table~\ref{tab:acd-results} we compare official results when generating the 3 TIAD test language pairs from scratch. Details on the evaluation procedure are available on this edition's webpage \footnote{https://tiad2021.unizar.es/results.html}. Our submitted prediction set contained filtered translations only above confidence $0.5$ and thus we show the official results in the confidence range $[0.5, 1]$. We have also included OTIC results with in the confidence range $[0 to 1$] as a baseline. It is clear that while OTIC has slightly higher precision, it produces a much smaller prediction set.

In table~\ref{tab:En-Ca_Results} we compare OTIC, CD and ACD on the En-Ca generation validation task for more detailed insight. The results of OTIC are better than CD as both the En-Es and Es-Ca dictionaries are quite rich. ACD significantly improves the coverage and recall at a reasonable drop in precision compared to OTIC. We believe that above a certain accuracy, generating larger prediction sets is more important as predictions are likely to be human validated anyway. 

\begin{table}[t]
\centering
\resizebox{\textwidth}{!}{%
\begin{tabular}{|p{16mm}|p{15mm}|p{12mm}|p{10mm}|p{16mm}|}
\hline
\textbf{Threshold} & \textbf{Precision} & \textbf{Recall} & \textbf{F1} & \textbf{Coverage} \\  
\hline
0.5       & 0.75      & 0.53   & 0.61      & 0.75     \\ 
\hline
0.6       & 0.76      & 0.52   & 0.62      & 0.74     \\ 
\hline
0.7       & 0.78      & 0.51   & 0.62      & 0.73     \\ 
\hline
0.8       & 0.78      & 0.51   & 0.62      & 0.73     \\ 
\hline
0.9       & 0.80      & 0.49   & 0.61      & 0.71     \\ 
\hline
1         & 0.82      & 0.47   & 0.59      & 0.68     \\ 
\hline
\end{tabular}%

\quad

\begin{tabular}{|p{16mm}|p{15mm}|p{12mm}|p{10mm}|p{16mm}|}
\hline
\textbf{Threshold} & \textbf{Precision} & \textbf{Recall} & \textbf{F1} & \textbf{Coverage} \\  \hline
0         & 0.76      & 0.19   & 0.30      & 0.29     \\ \hline
0.1       & 0.76      & 0.19   & 0.30      & 0.29     \\ \hline
0.2       & 0.76      & 0.19   & 0.30      & 0.29     \\ \hline
0.3       & 0.76      & 0.19   & 0.29      & 0.29     \\ \hline
0.4       & 0.77      & 0.18   & 0.29      & 0.29     \\ \hline
0.5       & 0.78      & 0.18   & 0.29      & 0.28     \\ \hline
0.6       & 0.78      & 0.16   & 0.27      & 0.26     \\ \hline
0.7       & 0.81      & 0.12   & 0.21      & 0.20     \\ \hline
0.8       & 0.85      & 0.12   & 0.21      & 0.20     \\ \hline
0.9       & 0.85      & 0.12   & 0.21      & 0.19     \\ \hline
1         & 0.85      & 0.12   & 0.21      & 0.19     \\ \hline
\end{tabular}%
}

\caption{TIAD'21 official results at varying confidence thresholds for ACD (left) vs OTIC (right). }
\label{tab:acd-results}

\centering
\resizebox{0.6\textwidth}{!}{%
\begin{tabular}{|p{17mm}|p{17mm}|p{15mm}|p{17mm}|p{17mm}|}
\hline
\textbf{Metric} & \textbf{OTIC} & \textbf{CD} & \textbf{ACD}\\ \hline
\textbf{BWP}         & 97.56\%      & 85.25\%   & 86.26\%     \\ \hline
\textbf{BWR}       & 100\%      & 47.40\%   & 75.93\%     \\ \hline
\textbf{Recall}       & 30.99\%      & 24.60\%   & 39.40\% \\ \hline
\textbf{Covg.}       & 34.43\%      & 33.04\%   & 52.31\%     \\ \hline
\end{tabular}%
}

\caption{Comparison of OTIC, CD, and ACD on the En-Ca generation task. Note that as discussed in \cite{Goel2021} BWR depends on the input dataset, and merely reflects data efficiency when the input set differs. Thus while the BWR of CD and ACD are comparable as they have the same input data, OTIC's perfect BWR is incomparable. We also include traditional recall for a direct comparison between all three.}
\label{tab:En-Ca_Results}

\end{table}

 \titlespacing\section{0pt}{0pt plus 0pt minus 0pt}{6pt plus 0pt minus 0pt}
 
 \section{Conclusion}
In this paper, we showed a fundamental connection between the 2 most popular dictionary induction algorithms that don't need any additional resources except translation pairs: OTIC and Cycle Density. We proposed the Augmented Cycle Density method that enhances the cycle-based generation step in OTIC (Type-A translations) with the more sophisticated Cycle Density algorithm. With this, we beat the OTIC baseline for the first time in the history of the TIAD shared task. By subsuming both the CD and OTIC algorithms, ACD's performance improves with the availability of richer input translation graphs while retaining computational and data efficiency. This makes it a great fit for practical dictionary generation, and we hope to incorporate this functionality in the ApertiumBidixGen open-source tool soon. We hope the interactions of the Type-A and Type-B step in ACD with varying output tasks, input data and hyperparameter choices are studied in greater detail in the future. 

 \titlespacing\section{0pt}{6pt plus 0pt minus 0pt}{6pt plus 0pt minus 0pt}


\begin{thebibliography}{8}
\bibitem{Villegas2016}
M. Villegas, M. Melero, N. Bel and J. Gracia, Leveraging RDF graphs for crossing multiple bilingual dictionaries, in: Proceedings of the Tenth International Conference on Language Resources and Evaluation (LREC’16), 2016, pp. 868–876.

\bibitem{Goel2021}
S. Goel, J. Gracia, M. Forcada, Bilingual Dictionary Generation and Enrichment via Graph Exploration, Under Review in Semantic Web Journal, http://www.semantic-web-journal.net/content/bilingual-dictionary-generation-and-enrichment-graph-exploration (2021).

\bibitem{Tanaka1998}
K. Tanaka and K. Umemura, Construction of a Bilingual Dictionary Intermediated by a Third Language., in: Proc. of The 15th International Conference on Computational Linguistics
(COLING’94), 1994, pp. 297–303

\bibitem{Liu2018Homographs}
Frederick Liu, Han Lu, and Graham Neubig. Handling homographs in neural machine translation. In Proceedings of the 2018 Conference of the North American Chapter of the Association for Computational Linguistics: Human  Language Technologies,  Volume  1  (Long  Papers), pages 1336–1345, New Orleans, Louisiana, June 2018. Association for Computational Linguistics.

\bibitem{Arora2018Linear}
 Sanjeev Arora, Yuanzhi Li, Yingyu Liang, Tengyu Ma, and Andrej Risteski. Linear algebraic structure of  word  senses, with applications to polysemy. Transactions  of  the Association for Computational Linguistics, 6:483–495, 2018.
 
 \bibitem{Mausam2009}
 Mausam, S. Soderland, O. Etzioni, D. Weld, M. Skinner and J. Bilmes, Compiling a Massive, Multilingual Dictionary via Probabilistic Inference, in: Proc. of the Joint Conference of the 47th Annual Meeting of the ACL and the 4th International Joint Conference on Natural Language Processing of the AFNLP, Association for Computational Linguistics, Suntec, Singapore, 2009, pp. 262–270. https://www.aclweb.org/anthology/P09-1030.
 
 \bibitem{Gracia2018RDF}
 J. Gracia, M. Villegas, A. Gomez-Perez and N. Bel, The apertium bilingual dictionaries on the web of data, Semantic Web 9(2) (2018), 231–240.
 
 \bibitem{Gracia2020leveraging}
 J. Gracia, C. Fäth, M. Hartung, M. Ionov, J. Bosque-Gil, S. Veríssimo, C. Chiarcos and M. Orlikowski, Leveraging Linguistic Linked Data for Cross-Lingual Model Transfer in the Pharmaceutical Domain, in: Proc. of 19th International Semantic Web Conference (ISWC 2020), B. Fu and A. Polleres, eds, Springer, 2020, pp. 499–514. ISBN 978-3-030-62465-1.
 
 \bibitem{Cimiano2020LLD}
P. Cimiano, C. Chiarcos, J.P. McCrae and J. Gracia, Linguistic Linked Data, Springer International Publishing, 2020. ISBN 978-3-030-30224-5
\end{thebibliography}
\end{document}